\documentclass[runningheads]{llncs}
\usepackage{authblk}
\usepackage[table]{xcolor}
\usepackage{float}

\usepackage{threeparttable}

\usepackage{stfloats}
\usepackage{amssymb}
\usepackage[inline]{enumitem} 
\usepackage{bm}
\usepackage{mathtools}
\usepackage{multirow}
\usepackage{makecell}
\usepackage{bbm}
\usepackage{cite}
\usepackage{multicol}
\usepackage{blindtext}
\usepackage{wrapfig,lipsum,booktabs}

\usepackage{times}
\usepackage{latexsym}

\usepackage[T1]{fontenc}

\usepackage[utf8]{inputenc}
\usepackage{amsmath}
\usepackage{tabularx}
\usepackage{footnote}
\usepackage{tablefootnote}
\makesavenoteenv{tabular}
\usepackage{microtype}

\begin{document}
\title{Topics in the Haystack: Extracting and Evaluating Topics beyond Coherence}

\author{Anton Thielmann\inst{1}\and
Quentin Seifert\inst{2}\and
Arik Reuter\inst{1}\and
Elisabeth Bergherr\inst{2}\and
Benjamin Säfken\inst{1}}

\authorrunning{A. Thielmann et al.}

\institute{Chair of Data Science and Applied Statistics, TU Clausthal, Germany \and
Chair of Spatial Data Science and Statistical Learning,  University of Göttingen, Germany
\email{anton.thielmann@tu-clausthal.de}}

\maketitle

\begin{abstract}

Extracting and identifying latent topics in large text corpora has gained increasing importance in Natural Language Processing (NLP). Most models, whether probabilistic models similar to Latent Dirichlet Allocation (LDA) or neural topic models, follow the same underlying approach of topic interpretability and topic extraction. 
We propose a method that incorporates a deeper understanding of both sentence and document themes, and goes beyond simply analyzing word frequencies in the data. This allows our model to detect latent topics that may include uncommon words or neologisms, as well as words not present in the documents themselves.  
Additionally, we propose several new evaluation metrics based on intruder words and similarity measures in the semantic space. We present correlation coefficients with human identification of intruder words and achieve near-human level results at the word-intrusion task. We demonstrate the competitive performance of our method with a large benchmark study, and achieve superior results compared to state-of-the-art topic modeling and document clustering models.
\end{abstract}

\setcounter{page}{1}

\section{Introduction}
Identifying latent topics in large text corpora is a central task in Natural Language Processing (NLP). With the ever-growing availability of textual data in virtually all languages and about every possible topic, automated topic extraction is gaining increasing importance. Hence, the approaches are manifold. For almost all models, a topic is intuitively defined by a set of words with each word having a probability of occurrence for the given topic. Different topics can share words, and a document can be linked to more than one topic. Generative probabilistic models, such as probabilistic latent semantic analysis (PLSA) \cite{hofmann2001unsupervised} and Latent Dirichlet Allocation (LDA) \cite{blei2003latent}, are still widely used and inspired multiple adaptations as e.g. \cite{agarwal2010flda, blei2010nested, chien2018latent, ramage2009labeled, rosen2012author} all drawing heavily from word-co-occurrences. Due to its popularity and general good performance on benchmark datasets, the interpretation of a topic from LDA is seldomly challenged. Neural topic models, like e.g. \cite{dieng2020topic, wang2019atm}, further improve upon the existing methods by integrating \textit{word-embeddings} or variational autoencoders \cite{srivastava2017autoencoding} into the  modeling approach, but still heavily rely on the ideas from \cite{blei2003latent}. 

New methods that challenge the typical idea of topic modeling also integrate \textit{word-} and \textit{document-embeddings} \cite{angelov2020Top2Vec, grootendorst2022bertopic, sia2020tired}. However, improvement over the current state of the art is usually measured in terms of performance as determined by evaluation metrics on standard benchmark datasets. While older models were still evaluated using likelihood-based perplexity metrics \cite{lafferty2005correlated, larochelle2012neural, rosen2012author}, empirical results showed a negative correlation between perplexity based metrics and human evaluation of a topic model \cite{chang2009reading}. Additionally, Chang et al. \cite{chang2009reading} first introduced the idea of \textit{intruder words}. According to this idea, a topic is  considered \textit{coherent} or simply put, \textit{good}, if a randomly chosen word, not belonging to that topic, can clearly be identified by humans. As human evaluation of models is cost and time intensive, researchers used new evaluation methods that correlated with human evaluation \cite{lau2014machine, newman2010automatic}. Hoyle et al. \cite{hoyle2021automated} even found no contemporary model at all that used human feedback as a form of model evaluation. Newer models were hence evaluated using coherence scores \cite{angelov2020Top2Vec, dieng2020topic,  grootendorst2022bertopic, sia2020tired,  srivastava2017autoencoding}.
However, Hoyle et al. \cite{hoyle2021automated} found severe flaws in coherence scores. First, they find that coherence scores exaggerate differences between models and second, they validate the findings from  Bhatia et al. \cite{bhatia2017automatic} and find much lower Pearson correlations between automated coherence scores and human evaluation as compared to \cite{lau2014machine}. 

We identify two shortcomings in the current state-of-the-art in topic modelling. The first is the significant gap in validated automatic evaluation methods for topic models. The second stems from the continued reliance on evaluation methods based on word co-occurrences and outdated definitions of topics from older models. Current methods rely on limited corpora from which the topic representations are created. However, integrating larger corpora into the modeling process can enhance topic quality by including contextually relevant words that were missing from the original corpus.

\subsubsection{Contributions} The contributions of this paper are hence twofold and can be summarized as follows:
\begin{itemize}
    \item We propose the Context Based Topic Model (CBTM) that, with only a few adaptations, integrates linguistic ideas into its modeling. Soft-clustering on the document level is integrated, such that P(document | topic) is modeled.
    \item We introduce  new topic modeling performance metrics. The validation of the proposed metrics is validated by demonstrating impressive correlations with human judgement.
    \item We conduct a benchmark study comparing the presented approach to state-of-the-art topic modeling and document clustering methods and outperform common benchmark models on both, coherence scores and the presented new metrics for topic evaluation.
\end{itemize}

The remainder of the paper is structured as follows: First, a short introduction into the used linguistic ideas and the definition of topics is presented. Second, the method of extracting latent topics from documents, incorporating the aforementioned definitions, is presented. Third, new evaluation metrics are introduced and validated by presenting correlations with human annotators. Fourth, the proposed model is applied to two common data sets and compared with state-of-the-art topic models. Finally, a discussion of the limitations as well as a conclusion is given in sections \ref{conclusion} and \ref{discussion}.

\section{On the Nature of Topics} \label{motivation}

\begin{figure}[h]
    \centering
    \includegraphics[scale=0.45]{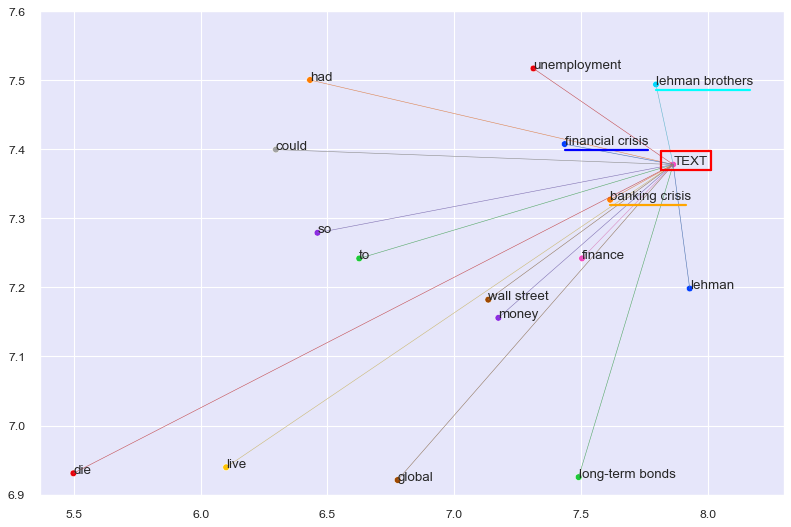}
    \caption{\footnotesize{The word best representing a sentence (or document) does not necessarily needs to be included in that text. The figure represents a New York Times headline from the financial crisis in 2009: ``\textbf{Lehman had to die so Global Finance could live}''. All words present in that text and additional words are mapped into a high dimensional feature space. The dimensions are reduced to visually demonstrate, that words not occurring in that sentence, e.g. \textbf{banking crisis} are better suited to summarize that sentence than words present in the sentence, e.g. \textbf{global}.}}
    \label{context}
\end{figure}

While there have been numerous approaches to extracting latent topics from large text corpora, little effort has been made in adapting those models to more refined definitions of a topic.
We propose a topic model that follows ideas from linguistic definitions of topics \cite{davison1982systematic, davison1984syntactic}. We present two ideas from linguistic theory in order to construct more humanly interpretable topics:
\begin{itemize}
    \item[\bf{i)}] A  word that most accurately expresses the topic of a document may not necessarily occur in that document.
    \item[\bf{ii)}] Only using nouns and noun phrases is more appropriate for representing understandable topics.
\end{itemize}

\noindent \textbf{i)} closely follows Guijarro \cite{guijarro2000towards}: "a topic is, above all, a textual category that is determined by the context and not by purely formal or structural aspects." Therefore, the topic of a document or even a sentence may go beyond the mere occurrence of all the words in that document. That is, a word that most accurately expresses the topic of a document may not necessarily occur in that document. We leverage a simple example from a New York Times headline to demonstrate that: 

\textbf{"Lehman had to die so Global Finance could live"} 

\noindent That sentence pertains to the financial crisis and the collapse of the Lehman Brothers bank, but neither phrase is explicitly mentioned. A bag-of-words model that only considers words present in the document corpus would not be able to accurately capture the document's topic. Contextually relevant words, even if not present in the document, can provide better representations.
Figure \ref{context} shows the described example. Comparing the cosine distance in a reduced embedding space between the complete embedded sentence (TEXT) and each embedded word demonstrates how words and phrases not occurring in that text can be a meaningful summary of that text. "Banking crisis" is a more meaningful representation of the sentence than e.g. "global" and lies closer to the text in the semantic space.
%

\noindent Common topic models, such as \cite{blei2003latent, dieng2020topic, bianchi-etal-2021-cross, srivastava2017autoencoding}, as well as document clustering methods, such as \cite{grootendorst2022bertopic, angelov2020Top2Vec, sia2020tired}, face a limitation in that they only consider words that appear in the reference corpus when generating topic representations. This limitation can lead to incorrect topic interpretations, as shown in the example above.
Through expanding the reference corpus and leveraging pre-trained embedding models, we make sure that "the indispensability of frame knowledge for understanding texts" \cite{beghtol1986bibliographic} is accounted for. 

\textbf{ii)} closely follows Beghto \cite{beghtol1986bibliographic}, after whom one of the features of generalized titles is the absence of verbal forms. Following the idea that a title is the highest macroproposition of a textual unit \cite{beghtol1986bibliographic}, we apply this idea to the construction of topics and hence propose to only consider nouns and noun phrases for the proposed method of topic extraction. 



\section{Methodology}
Let $\textup{V} =\left \{ w_1,~\ldots~, w_n   \right \}$ be the vocabulary of words and $\textup{D} =\left \{ d_1,~\ldots~, d_M   \right \}$ be a corpus, i.e. a collection of documents. Each document is a sequence of words $d_i = \left [ w_{i1}, \ldots, w_{in_i} \right ]$  where $w_{ij} \in V$ and $n_i$ denotes the length of document $d_i$. Further, let $\mathcal{D} = \left \{ \bm{\delta}_1,~\ldots~, \bm{\delta}_M   \right \}$ be the set of documents represented in the embedding space, such that $\bm{\delta}_i$ is the vector representation of $d_i$ and let $\mathcal{W} = \left \{ \bm{\omega}_1,~\ldots~, \bm{\omega}_n   \right \}$ be the vocabulary's representation in the same embedding space. Hence, each word $w_i$ in the embedding space represented as $\bm{\omega}_i  \in \mathbb{R}^{L}$ has the same dimensionality $L$ as a document vector $\bm{\delta}_i  \in \mathbb{R}^{L}$.
There are different representations of topics, but mostly a topic $t_k$ from a set of topics $ T = \left \{ t_1,~\ldots~, t_K   \right \}$ is represented as a discrete probability distribution over the vocabulary \cite{blei2003latent}, such that $t_k$ is often expressed as $(\phi_{k, 1}, \ldots, \phi_{k, n} )^T$ and $\sum_{i=1}^{n} \phi_{k, i}=1$ for every $k$\footnote{See table \ref{variable_list} in the Appendix for a complete variable and notation list}.

Based upon the idea expressed in section \ref{motivation}, we form clusters from the documents embeddings, $\mathcal{D}$ and subsequently extract topics, $t_k$, that represent these clusters best. Hence, after transforming the raw documents into document vectors, they are clustered. Due to the curse of dimensionality \cite{aggarwal2001surprising} we reduce the dimensions before clustering using UMAP \cite{mcinnes2018umap}, closely following \cite{angelov2020Top2Vec} and \cite{grootendorst2022bertopic}. However, we allow each document to belong to more than one cluster resulting in document topic matrices $\bm{\theta}$ and word topic matrices $\bm{\beta}$, similar to LDA \cite{blei2003latent}. The documents are clustered with a Gaussian mixture model \cite{reynolds2009gaussian}, as it not only allows for soft-clustering, but also has the advantage of optimizing hyperparameters via, for instance, the Akaike information criterion or the Bayesian information criterion. As a results, CBTM, in contrast to \cite{angelov2020Top2Vec, grootendorst2022bertopic, sia2020tired} offers not only word-topic distributions but also document-topic distributions.

\subsection{Topic Extraction} \label{extraction}
To find the words that best represent the corpus' topics, we first extract the centroids of the $k$ clusters, $\bm{\mu}_k  \in \mathbb{R}^{L}$, in the original embedding space. Second, we filter the given vocabulary for nouns and enhance this vocabulary by 
any specified external vocabulary of nouns, resulting in a new dictionary $\hat{V} = \left \{ w_1,~\ldots~, w_n, w_{n+1}, ~\ldots~, w_{n+z}\right \}$. The word vectors $\bm{\omega}_i$ closest to $\bm{\mu}_k$ in the embedding space, are the words that represent cluster $k$'s centroid best \cite{angelov2020Top2Vec}, where it could happen, that a word represents a topic ideally where $w \notin V$ but always $w \in \hat{V}$. 
To compute the words best representing a topic, we compute the cosine similarity between every word in $\hat{V}$ and all cluster centroids in the embedding space. For a single word $w$, its embedding $\bm{\omega}$ and a single cluster with centroid $\bm{\mu}$, we hence compute:

\begin{equation}
    sim(\bm{\omega}, \bm{\mu}) = \frac{\bm{\omega} \cdot \bm{\mu}}{\lVert \bm{\omega}\rVert \Vert\bm{\mu}\lVert}, 
\end{equation}

\noindent where $\bm{\omega} \cdot \bm{\mu}=\sum_{i=1}^{L} \omega_{i} \mu_{i}$ and $$\lVert\bm{\omega}\rVert  \lVert\bm{\mu}\rVert=\sqrt{\sum_{i=1}^{L} (\omega_i)^2}  \sqrt{\sum_{i=1}^{L} (\mu_i)^2}.$$ $L$ denotes the vectors dimension in the feature space which is identical for $\bm{\omega}$ and $\bm{\mu}$.

To avoid having words in a topic that are semantically overly similar, as e.g. \textit{economics} and \textit{economy}, each topic can be \textit{cleaned}. The cosine similarity between the top $Z$ words contained in a topic can be computed and all words that exceed a certain threshold, e.g. 0.85\footnote{The cosine similarity between the words "economy" and "economies", using the paraphrase-MiniLM-L6-v2 embedder \cite{reimers-2019-sentence-bert} is for instance 0.9.}, are removed in descending order of the similarity with the clusters centroid. An additional advantage of the corpus expansion is the possibility to model documents in one language, but create topics in a different language, when using a multi-language embedding model.

\section{Evaluation} \label{metrics}

Given the described approach, we are effectively losing any idea of co-occurence based coherence for model evaluation. The words best describing a cluster of documents or topic do not necessarily have to occur  together often in documents. In fact, a word capturing the topic of a single document optimally, does not necessarily have to be contained in that same document. Additionally, by enhancing the corpus, it might be possible that neologisms are the words best representing a topic. Imagine, e.g. a set of documents being equally about software and hardware issues. The neologism \textit{software-hardware} would be an understandable and reasonable word describing that topic, but would perform poorly in any word-co-occurence based evaluation measure.

\subsection{Evaluation Metrics}
\begin{figure}
    \centering
    \includegraphics[scale=0.35]{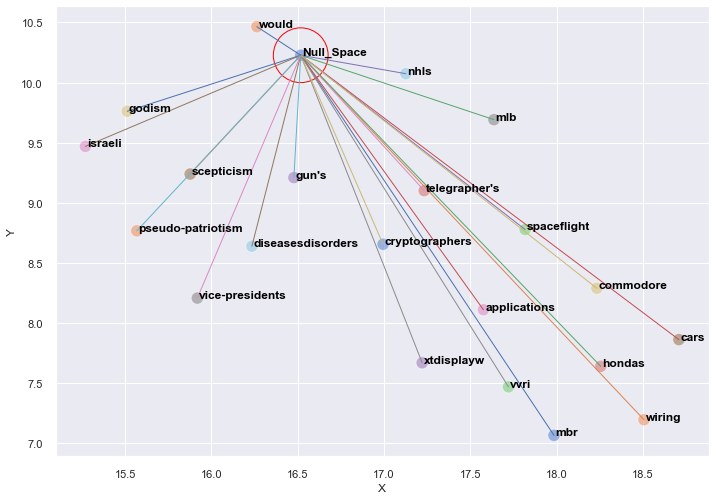}
    \caption{\footnotesize{The expressivity of a model is captured by averaging over the topics centroids cosine similarity to the null space, defined as the centroid of all embedded stopwords. For visualization the vector dimensions are heavily reduced, but the overall expressivity is still visualized. Due to the dimensionality reduction, the axes are just labelled "X" and "Y" respectively. The visualized topics are created from the 20 Newsgroups data set with the CBTM method and a single topic, "would", created with a LDA model. The topic's top word is annotated at the topic's position in the reduced embedding space.}}
    \label{eyxpressivity}
\end{figure}
For evaluation, we hence propose new, non word-co-occurence based measures and use existing measures leveraging word embeddings \cite{terragni2021word}. We validate the intruder based metrics by computing correlations with human annotations. 
\subsubsection{Topic Expressivity (EXPRS)}
First, we propose a novel measure inherently representing the meaningfulness of a topic. For that, we leverage stopwords, which widely recognized fulfill a grammatical purpose, but transport nothing about the meaning of a document \cite{salton1989automatic, wilbur1992automatic}. Hence, we compute the vector embeddings of all stopwords and calculate a centroid embedding. Subsequently, we compute the cosine similarity between a topic centroid and the stopword centroid (see Figure \ref{eyxpressivity}).

The weighted topic vector centroid, $\bm{\gamma}_k$, is computed by taking the top $Z$ words and normalizing their weights, such that $\sum_{i=1}^{Z} \phi_{k,i} =1$. The complete vector is hence computed as $\bm{\gamma}_k = \frac{1}{Z}\sum_{i=1}^{Z} \phi_{k,i}\bm{\omega_i}$ and the overall metric, which we call the models \textit{expressivity}, where we sum over all $K$ topics is defined as:
\begin{equation}
    EXPRS(\bm{\bm{\gamma}}, \bm{\psi}) = \frac{1}{K} \sum_{i=1}^{K} sim(\bm{\gamma}_i, \bm{\psi})
\end{equation}

\noindent with $\bm{\psi}$ being the centroid vector representation of all \textit{stopwords}. Note, that $\bm{\gamma}_i \neq \bm{\mu}_i$, as $\bm{\mu}_i$ is the centroid of the document cluster and $\bm{\gamma}_i$ is the centroid of topic $t_i$.

\subsubsection{Embedding Coherence (COH)}
\begin{figure}
    \centering
    \includegraphics[scale=0.35]{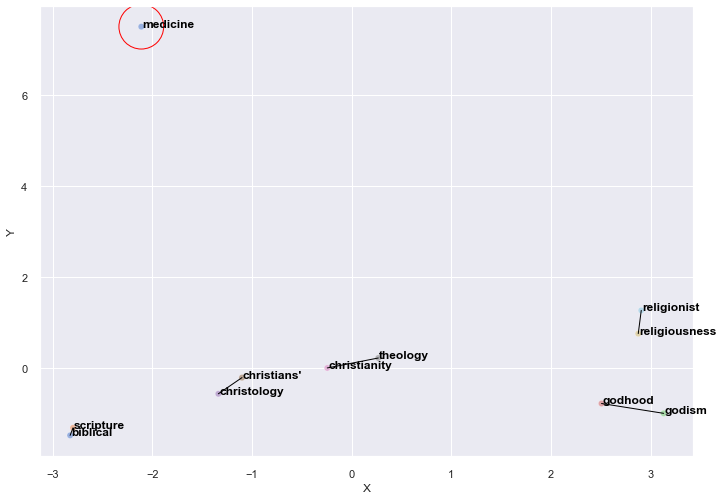}
    \caption{\footnotesize{The intruder word detection in the embedding space. A topic, covering "religion", and an intruder word, "medicine" are plotted with heavily reduced dimensions, using a PCA. The intruder word clearly separates from the otherwise coherent topic, even in a two-dimensional space. Due to the dimension reduction, the axis are just labelled with "X" and "Y" respectively. The topic is again created with the CBTM method on the 20 Newsgroups data set.}}
    \label{intruder_words}
\end{figure}
A measure, generally introduced by Aletras and Stevenson \cite{aletras2013evaluating} and reformulated by Fang et al. \cite{fang2016using} resembling classical coherence scores, is constructed by computing the similarity between the top $Z$ words in a topic. While Aletras and Stevenson \cite{aletras2013evaluating} compute the word vectors using word co-occurrences we follow Fang et al. \cite{fang2016using} and use the created word-embeddings. In contrast to classical coherence, we compute the similarity between every top-Z word in the topic and do not implement a sliding-window approach. Hence, for $Z$ words, we sum over $\frac{Z(Z-1)}{2}$ cosine similarities:
\begin{equation}
    COH(t_k) = \sum_{i=1}^{Z-1} \sum_{j = i + 1}^{Z} sim(\bm{\omega}_i, \bm{\omega}_j),
\end{equation}

\noindent where the overall average coherence of a model is hence computed as:
$$ \frac{2}{K (Z-1)Z} \sum_{k=1}^{K} COH(t_k).$$

\subsubsection{Word embedding-based Weighted Sum Similarity (WESS)}
A metric representing the diversity or the similarity between the topics of a topic model was introduced by \cite{terragni2021word} as the Word embedding-based Weighted Sum Similarity and is slightly adjusted for comparing models with a different number of topics as:

\begin{equation}
    WESS(T) = \frac{(K-1)K}{2}\sum_{i=1}^{K-1} \sum_{j =  i + 1}^{K} sim(\bm{\gamma}_i, \bm{\gamma}_j),
\end{equation}

\noindent where $\bm{\gamma}_i$ represents the weighted topic centroid for topic $i$. While this metric certainly captures the similarity between topics, it does also reflect the diversity of the model. Hence, if $WESS(T)$ is close to 1, the model would have created topics that are extremely similar to one another.

Additionally, we propose three different new metrics, leveraging the idea of intruder words \cite{chang2009reading} and similarly integrating an idea of topic diversity. First, a metric that is based upon unweighted topic centroids. 

\subsubsection{Intruder Shift (ISH)} 

Given the top $Z$ words from a topic, we calculate the topics unweighted centroid, denoted as $\tilde{\bm{\gamma}}_i$. Subsequently, we randomly select a word from that topic and replace it with a randomly selected word, from a randomly selected different topic. The centroid of the resulting words is again computed, denoted as $\hat{\bm{\gamma}}_i$. Given a coherent topic and generally diverse topics, one would expect a larger shift in the topics centroids. Therefore we calculate the \textit{intruder shift} of every topic and average over the number of topics:
\begin{equation}
    ISH(T) = \frac{1}{K} \sum_{i=1}^{K} sim(\tilde{\bm{\gamma}}_i, \hat{\bm{\gamma}}_i)
\end{equation}
\noindent Hence, one would expect a coherent and diverse topic model to have a lower $ISH$ score than an incoherent and non-diverse topic model.

\subsubsection{Intruder Accuracy (INT)}
The second intruder-word based metric follows the classical approach of identifying an intruder word more closely.
Given $Z$ top words of a topic, we again randomly select an intruder word from a randomly drawn topic. Subsequently, we calculate the cosine similarity for every possible pair of words within the set of the top $Z$ words. Then we calculate the cosine similarity of each top word and the intruder $\bm{\hat{\omega}}$.
Finally, our metric reports the fraction of top words to which the intruder has the least similar word embedding.

\begin{equation}
    \small{INT(t_k) = \frac{1}{Z}\sum_{i=1}^Z \mathbbm{1}(\forall j: sim(\bm{\omega}_i, \bm{\hat{\omega}}) < sim(\bm{\omega}_i, \bm{\omega}_j))}
\end{equation}

\noindent Hence we return the number of words from the set where the farthest word from them in the embedding space is the intruder word, divided by the number of words, $Z$, taken into account (See Figure \ref{intruder_words} for a visualization). 
\subsubsection{Average Intruder Similarity (ISIM)}
As a last metric, we propose the average cosine similarity between every word in a topic and an intruder word:

\begin{equation}
    ISIM(t_k) = \frac{1}{Z} \sum_{i=1}^{Z} sim(\bm{\omega}_i, \bm{\hat{\omega}})
\end{equation}

\noindent
To account for any induced randomness in the metrics $ISH$, $INT$ and $ISIM$ due to the random choice of a particular intruder from a particular topic, we propose to  calculate those metrics multiple times with differently chosen random intruder words and subsequently average the results. Hence, the robustness against the specific selection of intruder words is increased. 

\begin{table}
\centering
\small
    \caption{\small{Metric Evaluation: Accuracy and Pearson correlation with the reported \textit{true} (Intruder) and humanly selected (Human) intruder word from Chang et al. \cite{chang2009reading} for all models and all topics on the 20 Newsgroups dataset. As embedding models we consider the Paraphrase-MiniLM-L6-v2 model \cite{reimers-2019-sentence-bert}, the All-MiniLM-L12-v2 model \cite{wang2020minilm}, the All-mpnet-base-v2 model \cite{song2020mpnet}, the Multi-qa-mpnet-base-dot-v1 model \cite{song2020mpnet} and the All-distilroberta-v1 model \cite{liu2019roberta} as well as a word2vec model pre-trained on the GoogleNews corpus and a Glove model pre-trained on a Wikipedia corpus. The three best results for the human correlation and accuracy are marked in bold. 
    One can see that the metric evaluation for different embedding models produces impressive results, given the correlation between participants of 0.77. The paraphrase-MiniLM-L6-v2 performs best, considering $INT$ and $ISIM$, closely followed by the Glove model.}}
    \begin{tabular}{l|c c|c c}
         & \multicolumn{2}{c}{Accuracy} & \multicolumn{2}{c}{Correlation} \\
     Score & Intruder & \textbf{Human} & Intruder & \textbf{Human}\\
     \Xhline{3\arrayrulewidth}
    \multicolumn{5}{c}{Paraphrase-MiniLM-L6-v2} \\
    \Xhline{3\arrayrulewidth}
     
     $ISH$ & 0.613 &    0.512 & 0.526 & 0.492\\
    
     $INT$ &  0.722&  \textbf{0.622} &  0.775 &  \textbf{0.728}\\
    
     $ISIM$ & 0.810 &  \textbf{0.686} & 0.574 &  \textbf{0.539}\\
     \Xhline{3\arrayrulewidth}
     \multicolumn{5}{c}{Multi-qa-mpnet-base-dot-v1} \\
     \Xhline{3\arrayrulewidth}
     
     $ISH$ & 0.675 &    0.573 & 0.598 & 0.567 \\
    
     $INT$ & 0.700 &   \textbf{0.604} &  0.751 &  0.708\\
    
     $ISIM$ & 0.791 &   0.672 &  0.543 & 0.511\\
    \Xhline{3\arrayrulewidth} 
     \multicolumn{5}{c}{All-MiniLM-L12-v2} \\
     \Xhline{3\arrayrulewidth}
     
     $ISH$ &  0.766 &   \textbf{0.652} & 0.519 & \textbf{0.591}\\
    
     $INT$ & 0.677 &    0.58 &  0.723 & 0.687\\
    
     $ISIM$ & 0.766 &    0.652 & 0.519 &  0.490\\
    \Xhline{3\arrayrulewidth}
     \multicolumn{5}{c}{All-mpnet-base-v2} \\
     \Xhline{3\arrayrulewidth}
     
     $ISH$ &  0.763 &   \textbf{0.652} &  0.626 & \textbf{0.592}\\
    
     $INT$ & 0.661 &0.577 &  0.727 & 0.689\\
    
     $ISIM$ & 0.763 &    0.652 & 0.511 & 0.482\\
    \Xhline{3\arrayrulewidth}
    
    \multicolumn{5}{c}{All-distilroberta-v1} \\
     \Xhline{3\arrayrulewidth}
     
     $ISH$ & 0.766 &   \textbf{0.652} &  0.625 &  \textbf{0.592}\\

     $INT$ &  0.677 &   0.587 &  0.729 & 0.687\\
    
     $ISIM$ &  0.766 &    0.652 & 0.519 & 0.490\\
    \Xhline{3\arrayrulewidth}
    
    \multicolumn{5}{c}{word2vec GoogleNews} \\
     \Xhline{3\arrayrulewidth}
     
     $ISH$ & 0.413 &   0.335 &  0.338 &  0.302\\
    
     $INT$ &  0.719 &   0.603 &  0.774 & \textbf{0.715}\\
    
     $ISIM$ &  0.820 &    \textbf{0.684} & 0.554 & \textbf{0.506}\\
    \Xhline{3\arrayrulewidth}
    
    \multicolumn{5}{c}{Glove Wikipedia} \\
     \Xhline{3\arrayrulewidth}
     
     $ISH$ & 0.622 &	0.506 &	0.496 &	0.439\\
    
     $INT$ &  0.750 &	\textbf{0.634} &	0.786 &	\textbf{0.727}\\
    
     $ISIM$ &  0.808 &	\textbf{0.677} &	0.595 &	\textbf{0.549}\\
    \Xhline{3\arrayrulewidth}
    
    \end{tabular}
    
    \label{metric evaluation large table}
\end{table}

\subsection{Validation of Metrics}\label{validation}
To validate the intruder word based evaluation metrics we take the publicly available data from Chang et al. \cite{chang2009reading}. Similar to Lau et al. \cite{lau2014machine} we compute the metrics over all topics and all models provided in \cite{chang2009reading} for the 20 Newsgroups dataset. However, for clear interpretability, we reduce all words that include hyphens, due to the representations from \cite{chang2009reading}.
Hence, we compute the metrics for 7,004 topics in total. We compute the accuracy of the metrics in terms of the \textit{true} intruder and the humanly detected intruder for all metrics as well as the Pearson-\textit{r}. While the important measures are here the correlation with the human annotations, reporting the correlations with the \textit{true} intruder word ensures that the metrics are not inherently biased towards machine selection. For the accuracy, we consider a pre-selected or human-selected intruder to be correctly identified, if the score for this word is the lowest or highest, respectively, among all displayed top words. The results are shown in \textbf{Table \ref{metric evaluation large table}}. For all results it must be noted that the human answers have some ambiguity in them. As reported by Lau et al. \cite{lau2014machine}, the Pearson-\textit{r} between the human answers was 0.77. 

Hence, the results for $INT$ with a maximum correlation of 0.728 is highly credible and outperforms the reported correlations \cite{lau2014machine} for coherence evaluation metrics.
Interestingly, $ISIM$ performs best, when considering the accuracy for the \textit{true} intruder word, but significantly worse when considering the human selected word. We find that, independent of the chosen model, the newly introduced metrics strongly outperform the results reported by Lau et al. \cite{lau2014machine} at the topic-level with reported Pearson correlations of around $r = 0.6$.

\section{Results} \label{results}
To evaluate the proposed model, we compare the model results with different benchmark models. We also demonstrate the validity of our two hypotheses on corpus expansion and noun phrases stated in Section 2.

As comparison models, we use BERTopic \cite{grootendorst2022bertopic} and Top2Vec \cite{angelov2020Top2Vec} as closely related models and representatives of clustering based topic models, LDA \cite{blei2003latent} as a model not leveraging pre-trained embeddings, CTM \cite{bianchi-etal-2021-cross} as a generative probabilistic model leveraging pre-trained embeddings, a simple K-Means model - closely following the architecture from \cite{grootendorst2022bertopic}, but replacing HDBSCAN with a K-Means clustering approach, ETM \cite{dieng2020topic} leveraging word2vec \cite{le2014distributed} and NeuralLDA and ProdLDA \cite{srivastava2017autoencoding}. All models are fit using the OCTIS framework \cite{terragni2021octis}. Where applicable the same pre-trained embedding model as for CBTM, \textit{all-MiniLM-L6-v2} \cite{reimers-2019-sentence-bert} is used. Note, that we perform extensive hyperparameter tuning for all models except for CBTM. A detailed description of the benchmark models hyperparameters and the hyperparameter tuning can be found in the Appendix. As a corpus expanding the reference corpus in CBTM for topic extraction we use the Brown corpus taken from nltk \cite{bird2009natural}, which we also use for filtering the vocabulary for noun-phrases. We compute the proposed metrics from Section \ref{metrics} except for the $ISH$ metric due to its inferior performance on the intruder word detection task (Table \ref{metric evaluation large table}). Additionally, we compute normalized pointwise mutual information (NPMI) scores \cite{lau2014machine} with the input corpus as the reference corpus and Topic Diversity (WESS) and Word-embedding Pairwise Coherence scores (COHPW) using the OCTIS framework \cite{terragni2021octis}. All word-embedding based metrics are computed with the \textit{paraphrase-MiniLM-L6-v2} model \cite{reimers-2019-sentence-bert} due to the results from Table \ref{metric evaluation large table}, except for WESS and COHPW where we use OCTIS' default pre-trained word2vec \cite{le2014distributed} model\footnote{ The word2vec model is trained on the GoogleNews corpus. The number of top words, $Z$, taken into account for the metrics $EXPRS$, $COH$, $WESS$, $INT$ and $ISIM$ is 10. For $INT$ and $ISIM$, we randomly select an intruder word from a randomly selected topic 50 times and report the averages.}. 

\begin{table}
\centering
\small

    \caption{\small{Comparison of noun-based topic extraction vs. non-noun-based model extraction for the CBTM model. The reported metrics are averaged over the results for three datasets, the 20 Newsgroups dataset, the BBC News dataset and the M10 dataset. All datasets are taken from OCTIS. All models are fitted using the all-MiniLM-L6-v2 model \cite{reimers-2019-sentence-bert}. Given the results from Table \ref{metric evaluation large table}, \textit{paraphrase-MiniLM-L6-v2} is used for the embedding based evaluation metrics. We report the baseline metrics for a model not using an expanded corpus and using all word types and report the differences to that baseline. We find that especially expanding the reference corpus leads to better topics, represented by nearly all metrics. As expected, the NPMI coherence scores are considerably worse, when expanding the reference corpus. That is due to the fact, that we used the original corpus the models where fit on as the NPMI coherence reference corpus.
    Additionally, we find that only considering nouns for topic words, can increase the evaluation metrics, especially when we clean the topics.}}
    \begin{threeparttable}
    \resizebox{\textwidth}{!}{
    \begin{tabular}{p{1.3cm}|c c c|c c c|c c}
    & \multicolumn{3}{c}{Coherence Measures} &  \multicolumn{3}{c}{Diversity Measures} & 
     \multicolumn{2}{c}{Intruder Measures} \\[0.1cm]
    Model & NPMI ($\uparrow$) &  COHPW ($\uparrow$) & COH ($\uparrow$) & TOP DIV ($\uparrow$) & WESS ($\downarrow$) & EXPRS ($\downarrow$)  & ISIM ($\downarrow$) & INT ($\uparrow$) \\ [0.1cm]
    \Xhline{3\arrayrulewidth}
    
     CBTM\tnote{*}  &  0.016 & 0.430 & 0.427 & 0.783 & 0.377 & 0.459  &  0.184 &  0.719 \\ 
     
     \Xhline{3\arrayrulewidth}
     
     CBTM\tnote{+}  &  -0.757 &  -0.079 &  +0.07 & +0.08 & -0.069 &  -0.061  &  -0.019 & +0.085 \\
     
     CBTM\tnote{**}  &  -0.018 &  +0.011 &  -0.014 &  -0.013 & +0.016 &  +0.007  &  +0.004 & -0.031 \\
     
     CBTM\tnote{**+}  & -0.70 & -0.073 &  +0.011 & +0.052 & -0.046 &  -0.043  & -0.027 &  +0.050 \\
     
     \Xhline{3\arrayrulewidth} 
     \multicolumn{9}{c}{\textbf{Cleaned with Similarity Threshold of 0.85}} \\[0.2cm]
     \Xhline{3\arrayrulewidth}
     
     CBTM\tnote{*}  &  0.014 & 0.433 & 0.421 & 0.775 & 0.386 & 0.467  &  0.189 &  0.708 \\ 
     
     \Xhline{3\arrayrulewidth}
     
     CBTM\tnote{+}  &  -0.752 &  -0.095 &  +0.042 & +0.077 & -0.060 &  -0.055  &  -0.026 & +0.066 \\
     
     CBTM\tnote{**}  &  -0.016 &  +0.010 &  +0.013 &  +0.003 & -0.012 &  +0.004  &  $\pm$0 & +0.021\\
     
     CBTM\tnote{**+}  & -0.689 & -0.081 &  +0.032 & +0.055 & -0.048 &  -0.045  & -0.029 &  +0.045 \\
    
    \hline
    \end{tabular}
    }
     \begin{tablenotes}
      \small
      \begin{enumerate*}
      \item[*]Baseline \item[**]Only Nouns \item[$^+$]Expanded
      \end{enumerate*}
    \end{tablenotes}
    \end{threeparttable}
    \label{robustness_check}

\end{table}

To confirm our two hypotheses from Section 2 that expanding the reference corpus and only considering nouns for topic extraction can increase the topic quality, we perform several analyses. We compare the presented method with and without reference corpus expansion and with and without noun phrase filtering. The averaged results over 3 datasets can be seen in \textbf{Table \ref{robustness_check}}.
\subsubsection{Hypothesis I: Corpus Expansion}
 Our results confirm our hypothesis that expanding the reference corpus leads to creating better topics depicted by nearly all metrics. Unsurprisingly, we find that NPMI coherence scores, only using the reference corpus for computing the coherence are decreased when expanding the reference corpus during topic extraction. Additionally, we find that using a smaller pre-trained model for computing the metrics, as the leveraged word2vec \cite{le2014distributed} model for COHPW and WESS also shows a decrease in performance when expanding the reference corpus. That is presumably due to the smaller vocabulary size used in these models.
\subsubsection{Hypothesis II: Noun Phrases}
We find that the noun-based models perform worse than the models that consider all types of words and for the different embedding models used to construct the evaluation metrics. However, we find that when cleaning the topics the topic quality increases when using only nouns as compared to using all word types. Additionally we find that expanding the reference corpus and only considering nouns achieves better performance than no expansion and using all word types.

\subsubsection{Benchmarks}

\begin{table}
\centering
\small
        \caption{\small{Benchmark results on the 20 Newsgroups and Reuters datasets. All models are fit using the \textit{all-MiniLM-L6-v2} pre-trained embedding model \cite{reimers-2019-sentence-bert} where applicable. \textit{paraphrase-MiniLM-L6-v2} is used for the evaluation metrics ISIM, INT, TOP DIV and EXPRS. For the metrics available in OCTIS we use the default embeddings which are pre-trained word2vec embeddings on the Google News corpus. Extensive hyperparameter tuning is performed for the comparison models (see Appendix). All models, except BERTopic and Top2Vec, are fit with a pre-specified number of 20 or 90 topics respectively. BERTopic and Top2Vec detect the \textit{optimal} number of topics automatically, hence we fit the model as intended by the authors. However, we additionally fit a K-Means model using the class based tf-idf topic extraction method from BERTopic with 20 and 90 topics respectively and hierarchically reduce the number of topics in Top2Vec.}}
    \begin{threeparttable}
    \resizebox{\textwidth}{!}{
    \begin{tabular}{p{1.45cm}|c c c|c c c|c c}
    & \multicolumn{3}{c}{Coherence Measures} &  \multicolumn{3}{c}{Diversity Measures} & 
     \multicolumn{2}{c}{Intruder Measures} \\[0.1cm]
    Model & NPMI ($\uparrow$) &  COHPW ($\uparrow$) & COH ($\uparrow$) & TOP DIV ($\uparrow$) & WESS ($\downarrow$) & EXPRS ($\downarrow$)  & ISIM ($\downarrow$) & INT ($\uparrow$)\\ [0.2cm]
    \Xhline{3\arrayrulewidth}\\
    & \multicolumn{8}{c}{\textbf{20 Newsgroups}} \\ \\
    \Xhline{3\arrayrulewidth}

	K-Means	                    & 0.080                         &	0.081  &  0.289 & 0.312 & 0.920 & 0.466  & 0.138 & 0.414 \\
	BERTopic\tnote{$\dagger$}	& 0.033                         &	0.039  &  0.244 & 0.362 & 0.607 & 0.499  & \cellcolor{green!30}0.151 & 0.280 \\
	Top2Vec\tnote{$\dagger$}	& \cellcolor{green!90}0.164     &	0.080  &  0.341 & 0.370 & 0.288 & 0.472 & 0.156 & 0.513  \\
	Top2Vec	                    & \cellcolor{green!60}0.158     &	0.100  &  0.384 & 0.346 & 0.825 & \cellcolor{green!30}0.442 & 0.152 & 0.654  \\
	LDA	                        & -0.141                        &	0.031  &  0.260 & 0.281 & 0.875 & 0.447 & 0.181 & 0.275  \\
	ProdLDA	                    & -0.003                        &	0.064  &  0.247 & 0.344 & 0.835 & 0.518 & 0.157 & 0.243 \\
	NeuralLDA	                & -0.187                        &	0.011  &  0.210 & 0.590 & 0.820 & 0.685 & 0.193 & 0.131 \\
	ETM	                        & -0.514                        &	0.038  &  0.274 & 0.634 & \cellcolor{green!30}0.265 & 0.695 & 0.259 & 0.197  \\
	CTM	                        & -0.069                        &	0.027  &  0.251 & 0.360 & 0.725 & 0.533 & 0.167 & 0.301 \\[0.1cm]
	\Xhline{3\arrayrulewidth}
	CBTM\tnote{$^{+}$}	        & -0.893                        &	\cellcolor{green!60}0.364  &  \cellcolor{green!90}0.523 & \cellcolor{green!90}0.925 & \cellcolor{green!90}0.234 & \cellcolor{green!90}0.368  & \cellcolor{green!60}0.130 & \cellcolor{green!90}0.886  \\
	CBTM\tnote{*}	            & \cellcolor{green!30}0.156     &	\cellcolor{green!90}0.443  &  \cellcolor{green!30}0.414 & \cellcolor{green!30}0.775 & 0.352 & 0.460  & 0.171 & \cellcolor{green!30}0.742  \\
	CBTM\tnote{*$^{+}$}	        & -0.807                        &	\cellcolor{green!30}0.342  &  \cellcolor{green!60}0.460 & \cellcolor{green!60}0.885 & \cellcolor{green!60}0.256 & \cellcolor{green!60}0.380  & \cellcolor{green!90}0.126 & \cellcolor{green!60}0.832  \\
	\Xhline{3\arrayrulewidth}\\
    & \multicolumn{8}{c}{\textbf{Reuters}} \\ \\
    \Xhline{3\arrayrulewidth}

	K-Means	                    & \cellcolor{green!90}-0.139 & 0.042 & 0.209 & 0.441 & 0.578 & 0.531 & \cellcolor{green!90}0.151 & 0.179 \\ 
	BERTopic\tnote{$\dagger$}	& -0.158 & 0.039 & 0.202 & 0.475 & 0.584 & 0.556 & \cellcolor{green!60}0.152 & 0.167 \\ 
	Top2Vec\tnote{$\dagger$}	& \cellcolor{green!30}-0.240 & 0.067 & 0.340 & 0.504 & \cellcolor{green!60}0.159 & 0.407 & 0.206 & 0.304 \\ 
	Top2Vec                     & \cellcolor{green!60}-0.168 & 0.075 & 0.367 & 0.505 & \cellcolor{green!30}0.271 & 0.388 & 0.216 & 0.376 \\ 
	LDA	                        & -0.822 & 0.025 & 0.387 & 0.533 & 0.394 & 0.660 & 0.364 & 0.172 \\ 
	ProdLDA	                    & -0.650 & 0.005 & 0.256 & 0.441 & 0.299 & 0.573 & 0.203 & 0.197 \\ 
	NeuralLDA	                & -0.446 & 0.013 & 0.209 & \cellcolor{green!60}0.645 & 0.920 & 0.733 & 0.196 & 0.129 \\ 
	ETM	                        & -0.920 & 0.008 & 0.486 & \cellcolor{green!90}0.676 & \cellcolor{green!90}0.096 & 0.671 & 0.467 & 0.190 \\ 
	CTM	                        & -0.602 & 0.012 & 0.285 & 0.441 & 0.362 & 0.617 & 0.237 & 0.209 \\ 
	[0.1cm]
	\Xhline{3\arrayrulewidth}
	CBTM\tnote{$^{+}$}	        & -0.581 & \cellcolor{green!30}0.167 & \cellcolor{green!90}0.489 & \cellcolor{green!30}0.539 & 0.316 & \cellcolor{green!90}0.320 & \cellcolor{green!30}0.172 & \cellcolor{green!90}0.695\\
	CBTM\tnote{*}	            & -0.252 & \cellcolor{green!90}0.198 & \cellcolor{green!30}0.421 & 0.458 & 0.365 & \cellcolor{green!30}0.344 & 0.176 & \cellcolor{green!30}0.610\\
	CBTM\tnote{*$^{+}$}	        & -0.252 & \cellcolor{green!60}0.179 & \cellcolor{green!60}0.431 & 0.483 & 0.356 & \cellcolor{green!60}0.339 & \cellcolor{green!30}0.172 & \cellcolor{green!60}0.643\\
 
    \end{tabular}
    }
     \begin{tablenotes}
      \small
      \begin{enumerate*}
      \item[$\dagger$] HDBSCAN results with > 20 or 90 topics respectively \\
      \item[*]Only Nouns \\
      \item[$^{+}$]Expanded topic corpus
      \end{enumerate*}
    \end{tablenotes}
    \end{threeparttable}
    \label{20NG_results}
\end{table}
For comparing CBTM with other models we use two standard benchmark datasets, 20 Newsgroups and Reuters \cite{lewis1997Reuters} as shown in \textbf{Table \ref{20NG_results}}. We fix the number of topics to the \textit{true} number of topics of 20 and 90, respectively (see Appendix for additional benchmarks on two further datasets). CBTM outperforms all models, concerning \textit{INT}, \textit{COH} and \textit{COHPW} for both datasets for all configurations. Additionally, CBTM performs well on topic diversity for the 20 Newsgroups dataset and \textit{EXPRS} for both datasets. Interestingly, it also performs very well concerning classical NPMI coherence scores for the 20 Newsgroups dataset when not expanding the reference corpus. As expected, the models closely related to CBTM perform also well on both datasets. However, while Top2Vec, BERTopic and the used K-Means model are closely related to the proposed CBTM, CBTM achieves much better results concerning all metrics. Interestingly, CTM performs very well on smaller datasets (see supplemental material for additional benchmarks). Additionally, our results do not confirm that models that use a hard clustering approach perform considerably worse for a multi-label dataset (Reuters) as compared to models that integrate soft-clustering (see e.g. CTM/ETM vs Top2Vec/BERTopic results).

\section{Conclusion} \label{conclusion}
We develop a novel model for topic extraction beyond the mere occurrence of words in the reference corpus. We are able to show that expanding the reference corpus improves model performance. Additionally, we can confirm, that restricting the word types for topic extraction by only considering nouns can also lead to improved topic quality, under certain conditions. CBTM outperforms commonly used state-of-the-art topic models on multiple benchmark datasets, even in cases where the comparison models underwent extensive hyperparameter tuning while no hyperparameter tuning was performed for CBTM (see supplemental material for details on the hyperparameter tuning).

Given that almost all newly introduced topic models are evaluated automatically \cite{hoyle2021automated}, automatic evaluation metrics are of outmost importance. Hoyle et al. \cite{hoyle2021automated} even postulated that automatic topic model evaluation is broken, as the current used metrics have overall low correlations with human judgement of topic quality. We present multiple novel evaluation metrics closely following state of the art human evaluation of topic model quality and achieve great correlations with human evaluation. We greatly improve upon the correlation with human evaluation compared to the currently most often used metric, NPMI, achieving correlations of around $r = 0.73$ compared to NPMI correlations of $r = 0.63$.
The proposed approach of using word embeddings and cosine similarity achieves impressive results given the overall lower agreement between human responses (Pearson-$r$=0.77). 

Additionally, we introduce a novel evaluation metric, based upon the centroid cluster of \textit{stopwords} in the embedding space. Given the approach of enhancing the reference corpus, the described model might be especially useful when evaluating short texts or identifying sparsely represented topics in a corpus \cite{thielmann2021one, thielmann2021unsupervised}. Through the inherent sparsity of the data, the words best describing a topic might not be included in the reference corpus and an enhancement could thus greatly improve the creation of topics.

\section{Limitations} \label{discussion}
Automated evaluation of topic model quality is inherently difficult. That difficulty is considerably increased by the fact there is no gold standard or even a ground truth for the quality of a topic. Chang et al. \cite{chang2009reading} introduced the reasonable approach of evaluating the coherence of a set of words with intruder-words. However, one cannot expect 100\% agreement between people when it comes to judging whether a word is an intruder word in a topic. The proposed evaluation metrics achieve impressive results with human annotations, they cannot, however, reflect human ambiguity or extreme subtlety in perceived topic quality. Additionally, as all evaluation metrics based upon human evaluation and hence experimental results achieved with human participants, the metrics might reflect a selection bias (WEIRD) \cite{henrich2010most}.
Further embedding models could be evaluated and tested and larger human evaluation studies could be conducted.

Recent findings about the dominance of certain dimensions in transformer embeddings \cite{timkey2021all} suggest an inherent bias in transformer embeddings that could negatively affect similarity measures in the semantic space. Our results do not suggest that such a bias negatively influences the modeling results, however, this study does not look into the dimensionality effects which could be the topic of further research.
 
Moreover, the creation of transformer models solely for the purpose of topic extraction that emphasize, for example, the beginnings of phrases due to their increased importance to the underlying topics of a subsubsection \cite{kieras1980initial, kieras1981topicalization} could greatly improve upon the existing methods.

\newpage
\bibliography{bib.bib}
\bibliographystyle{splncs04}

\newpage
\appendix
\section{Supplemental Methodology}
To make reading easier, we provide a full notation list. All used variables and their notation can be found here.

\begin{table}[h]
\small
  \caption{Variable list}
  
      \begin{tabular*}{\linewidth}{@{\extracolsep{\fill}}*8l@{}} 
$V$                     &   Vocabulary                                      \\
$D$                     &   Corpus                                          \\
$M$                     &   Number of documents in the corpus               \\
$d_i$                   &   Document $i$                                    \\
$w_i$                   &   Word $i$ in V                                   \\
$\bm{\omega}_i$         &   Word $i$ represented in the embedding space     \\
$\bm{\delta}_i$         &   Document $i$ represented in the embedding space \\

$\hat{\bm{\delta}_i}$ & $d_i$ represented in the reduced embedding space \\
$t_k$                   &   Topic $k$                                       \\
$T$                     &   Set of topics                                   \\
$\phi_{k,i}$            &   Probability of word $i$ in topic $k$            \\
$\bm{\gamma}_k$         &   Topic centroid vector of topic $k$              \\
$\bm{\mu}_k$            &   Mean of document cluster k                      \\
$\bm{\theta}$           &   Document cluster/topic matrix                   \\
$\bm{\beta}$            &   Word cluster/topic matrix                       \\
$\psi$                  &   Null Space/centroid of all stopwords            \\

    \end{tabular*}

    \label{variable_list}
\end{table}

All modeling steps from the proposed method are presented here in extensive form. 
First, the target corpus should be embedded. This can be done, either using contextualized transformer embeddings, as e.g. Bianchi et al. \cite{bianchi2020pre} showed that contextualized embeddings can improve topic quality. However, approaches as used by Sia et al. \cite{sia2020tired} where every word is embedded singularly and the documents are represented as centroid vectors of all occurring words are also possible. Second, the dimensions of the embedded documents, $\bm{\delta_i}$, are reduced due to the curse of dimensionality. Afterwards, the reduced embeddings, $\hat{\bm{\delta}_i}$, are clustered e.g. using GMM such that soft clustering is possible. The centroids for each document cluster, $\bm{\mu}_k$, are computed. Next, the corpus is filtered for nouns and all nouns present in the corpus supplemented by all nouns present in an expansion corpus are embedded. Note, that here the same embedding procedure must be chosen as for the documents ( see e.g. \cite{angelov2020Top2Vec, grootendorst2022bertopic}). Then, the similarity between all candidate words and all document cluster centroids is computed. Based on the candidate embeddings and the similarity to the document clusters $\bm{\mu}_k$, the topic centroids $\bm{\gamma}_k$ are computed and similar to LDA, we get a document topic matrix, $\bm{\theta}$, and a word topic matrix, $\bm{\beta}$. Last, a cleaning step can be performed to remove overly similar words from the topics.

\section{Human Topic Evaluation}
As automated evaluation of topic model quality is inherently difficult, creating great questionnaires and adequately operationalizing what researchers are interested in is adamantly important.
Lund et al. \cite{lund2019automatic} introduced a topic-word matching task, weighting and selecting answers from participants that have a high confidence and performed well on test questions. Choosing that approach reduces ambiguity in answers, but also induces a bias towards highly confident participants and neglects the subtle differences in perceived quality from humans.
\cite{newman2010automatic}, chose a straight-forward approach of letting humans rate the created topics quality. Choosing a 3-point scale for model evaluation, however, can induce unreliability of responses \cite{krosnick2018questionnaire}. \cite{bhatia2017automatic, bhatia2020topic} introduce a document-level topic model evaluation leveraging the intruder-topic task, also introduced in Chang et al. \cite{chang2009reading}. However, for direct annotation they also resort to a 3-point ordinal scale. Clark et al. \cite{clark2021all} even question human judgement all together; however, the used questionnaire design not only does not provide a midpoint but additionally can strongly induce a bias in preference due to a highly biasing follow up question \cite{clark2021all} (See e.g. \cite{lehman1992focus}).

\subsection{Additional Benchmark Results}
In addition to the 20 Newsgroups and Reuters dataset, we fit all models on the M10 and BBC News datasets. Both datasets are taken from OCTIS \cite{terragni2021octis}. CBTM again outperforms most other models on nearly all metrics. Interestingly, CTM achieves good results for the BBC News dataset, which is comparably small with <2.000 documents. For the M10 dataset, which is comprised of scientific papers and hence a more \textit{difficult} dataset, we find that topic expansion strongly improves the model performance.

\begin{table*}[htb]
\centering
\small
     \caption{\small{Benchmark results on the M10 dataset. All models are fit using the \textit{all-MiniLM-L6-v2} pre-trained embedding model \cite{reimers-2019-sentence-bert} where applicable. \textit{paraphrase-MiniLM-L6-v2} is used for the evaluation metrics ISIM, INT, TOP DIV and EXPRS. For the metrics available in OCTIS we use the default embeddings which are pre-trained word2vec embeddings on the Google News corpus. Extensive Hyperparameter tuning is performed for the comparison models (See Appendix). All models, except BERTopic and Top2Vec, are fit with a pre-specified number of 10 topics. BERTopic and Top2Vec detect the \textit{optimal} number of topics automatically, hence we fit the model as intended by the authors. However, we additionally fit a KMeans model using the class based tf-idf topic extraction method from BERTopic with 10 topics and hierarchically reduce the number of topics in Top2Vec.}}
     
    \begin{threeparttable}
    
    \resizebox{\textwidth}{!}{
    \begin{tabular}{p{1.45cm}|c c c|c c c|c c}
    & \multicolumn{3}{c}{Coherence Measures} &  \multicolumn{3}{c}{Diversity Measures} & 
     \multicolumn{2}{c}{Intruder Measures} \\[0.1cm]
    Model & NPMI ($\uparrow$) &  COHPW ($\uparrow$) & COH ($\uparrow$) & TOP DIV ($\uparrow$) & WESS ($\downarrow$) & EXPRS ($\downarrow$)  & ISIM ($\downarrow$) & INT ($\uparrow$)\\ [0.1cm]
    \Xhline{3\arrayrulewidth}

Kmeans	                  & \cellcolor{green!90}-0.108 & 0.063 & 0.254  & \cellcolor{green!30}0.940 & 0.354 & 0.458 &  \cellcolor{green!90}0.149 & 0.320\\
BERTopic\tnote{$\dagger$} & -0.318 & 0.056 & 0.231  & 0.628 & 0.424 & 0.514 &  \cellcolor{green!60}0.165 & 0.219\\
Top2Vec\tnote{$\dagger$}  & -0.345 & 0.083 & 0.315  & 0.060 & 0.547 & 0.478 & 0.220 & 0.326\\
TOP2Vec                   & -0.270 & 0.100 & 0.335  & 0.780 & 0.496 & 0.454 & 0.198 & 0.484\\
LDA	                      & \cellcolor{green!60}-0.176 & 0.035 & 0.244  & 0.830 & 0.330 & 0.440 & 0.208 & 0.177\\
ProdLDA	                  & -0.251 & 0.074 & 0.222  & \cellcolor{green!90}0.970 & 0.425 & 0.508 &  0.170 & 0.220\\
NeuralLDA	              & -0.571 & 0.030 & 0.186  & 0.373 & 0.582 & 0.581 & 0.185 & 0.118\\
ETM	                      & \cellcolor{green!30}-0.204 & 0.044 & 0.255  & 0.330 & 0.591 & 0.500 & 0.268 & 0.151\\
CTM	                      & -0.322 & 0.060 & 0.239  & \cellcolor{green!60}0.950 & \cellcolor{green!90}0.247 &  \cellcolor{green!90}0.353 & 0.172 & 0.271\\[0.1cm]
	\Xhline{3\arrayrulewidth}
	CBTM\tnote{$^{+}$}	       & -0.8411 & \cellcolor{green!30}0.338 & \cellcolor{green!90}0.512 & 0.855 & \cellcolor{green!60}0.322  &  \cellcolor{green!60}0.383 & 0.179 &	 \cellcolor{green!90}0.827\\
	CBTM\tnote{*}	           & -0.5762 & \cellcolor{green!90}0.441 & \cellcolor{green!30}0.419 & 0.770 & 0.394  & 0.420 & 0.193 &	 \cellcolor{green!30}0.719\\
	CBTM\tnote{*$^{+}$}	       & -0.8033 & \cellcolor{green!60}0.358 & \cellcolor{green!60}0.451 & 0.825 & \cellcolor{green!30}0.339  &  \cellcolor{green!30}0.395 &  \cellcolor{green!30}0.166 &	 \cellcolor{green!60}0.818\\
    \end{tabular}
    }
     \begin{tablenotes}
      \small
      \begin{enumerate*}
      \item[$\dagger$] HDBSCAN results with > 10 topics \item[*] Only Nouns \item[$^{+}$]Expanded topic corpus
      \end{enumerate*}
    \end{tablenotes}
    \end{threeparttable}
       
    \label{M10_results}
\end{table*}

 \begin{table*}[htb]
\centering
\small
\caption{\small{Benchmark results on the BBC News dataset. All models are fit using the \textit{all-MiniLM-L6-v2} pre-trained embedding model \cite{reimers-2019-sentence-bert} where applicable. \textit{paraphrase-MiniLM-L6-v2} is used for the evaluation metrics ISIM, INT, TOP DIV and EXPRS. For the metrics available in OCTIS we use the default embeddings which are pre-trained word2vec embeddings on the Google News corpus. Extensive Hyperparameter tuning is performed for the comparison models (See Appendix). All models, except BERTopic and Top2Vec, are fit with a pre-specified number of 10 topics. BERTopic and Top2Vec detect the \textit{optimal} number of topics automatically, hence we fit the model as intended by the authors. However, we additionally fit a KMeans model using the class based tf-idf topic extraction method from BERTopic with 5 topics and hierarchically reduce the number of topics in Top2Vec.}}
    \label{M10_results}
    \begin{threeparttable}
    \resizebox{\textwidth}{!}{
    \begin{tabular}{p{1.45cm}|c c c|c c c|c c}
    & \multicolumn{3}{c}{Coherence Measures} &  \multicolumn{3}{c}{Diversity Measures} & 
     \multicolumn{2}{c}{Intruder Measures} \\[0.1cm]
    Model & NPMI ($\uparrow$) &  COHPW ($\uparrow$) & COH ($\uparrow$) & TOP DIV ($\uparrow$) & WESS ($\downarrow$) & EXPRS ($\downarrow$)  & ISIM ($\downarrow$) & INT ($\uparrow$)\\ [0.1cm]
    \Xhline{3\arrayrulewidth}

Kmeans	                  &  -0.868 & 0.088 & 0.333 & \cellcolor{green!90}1.000 & \cellcolor{green!60}0.297 & 0.490  & \cellcolor{green!90}0.139 & \cellcolor{green!60}0.667 \\
BERTopic\tnote{$\dagger$} &  -0.307 & 0.053 & 0.232 & 0.623 & 0.423 & 0.513  & \cellcolor{green!30}0.166 & 0.218 \\
Top2Vec\tnote{$\dagger$}  &  -0.339 & 0.082 & 0.314 & 0.059 & 0.542 & 0.477  & 0.218 & 0.329 \\
TOP2Vec                   &  -0.324 & 0.097 & 0.334 & 0.920 & 0.419 & \cellcolor{green!60}0.435  & 0.173 & 0.528 \\
LDA	                      &  \cellcolor{green!60}-0.150 & 0.029 & 0.208 & 0.840 & 0.447 & 0.480  & 0.202 & 0.098 \\
ProdLDA	                  &  -0.290 & 0.050 & 0.212 & 0.960 & 0.484 & 0.541  & 0.171 & 0.199 \\
NeuralLDA	              &  -0.460 & 0.077 & 0.190 & \cellcolor{green!90}1.000 & 0.574 & 0.558  & 0.170 & 0.136 \\
ETM	                      &  \cellcolor{green!30}-0.184 & 0.043 & 0.249 & 0.600 & 0.510 & 0.489  & 0.252 & 0.182 \\
CTM	                      &  -0.299 & 0.050 & 0.232 & \cellcolor{green!90}1.000 & \cellcolor{green!90}0.236 & \cellcolor{green!90}0.369  & \cellcolor{green!60}0.148 & 0.241 \\[0.1cm]
	\Xhline{3\arrayrulewidth}
CBTM\tnote{$^{+}$}	       & -0.851	 & \cellcolor{green!30}0.351 & \cellcolor{green!90}0.456 & 0.810 &	\cellcolor{green!30}0.368 & \cellcolor{green!30}0.444 & 0.186 & \cellcolor{green!90}0.701 \\
CBTM\tnote{*}	           & \cellcolor{green!90}0.055  & \cellcolor{green!90}0.440 & \cellcolor{green!30}0.402 & 0.765 &	0.433 & 0.518 & 0.202 & 0.602 \\
CBTM\tnote{*$^{+}$}	       & -0.772	 & \cellcolor{green!60}0.373 & \cellcolor{green!60}0.403 & 0.795 &	0.397 & 0.474 & 0.181 & \cellcolor{green!30}0.656 \\

    \end{tabular}
    }
     \begin{tablenotes}
      \small
      \begin{enumerate*}
      \item[$\dagger$] HDBSCAN results with > 5 topics \item[*] Only Nouns \item[$^{+}$]Expanded topic corpus
      \end{enumerate*}
    \end{tablenotes}
    \end{threeparttable}
        
\end{table*}

\section{Experimental Setup}
For all tested models, we use the same pre-trained embedding model \textit{all-MiniLM-L6-v2}  \cite{reimers-2019-sentence-bert}, where applicable. NPMI Coherence scores are calculated as presented by \cite{lau2014machine}. For the best possible comparison, we use the same dimensionality reduction for CBTM as is used in Doc2Vec \cite{angelov2020Top2Vec} and BERTopic \cite{grootendorst2022bertopic}. Hence, we use Umap \cite{mcinnes2018umap} and reduce the dimensions to 5, explicitly using the same hyperparameters as done in the mentioned models. The same is done for the simple K-Means model.

\subsection{Hyperparameter Tuning}\label{hypertuning}
For CBTM we do not implement any form of hyperparameter tuning. Hence, the Gaussian Mixture Model is fit using scikit-learns default parameters. Hence the convergence threshhold for the Expectation Maximization (EM) Algorithm is 0.0001, each component has its own general covariance matrix and 1e-6 is added to the covariance diagonals for regularization purposes. The maximum number of iterations in the EM algorithm is set to 100 and K-Means is used to initialize the weights.

Hence, the results achieved by CBTM could be further optimized, by e.g. optimizing GMM with respect to the Bayesian- or Akaike Information Criterion. Additionally, the pre-trained embedding could be fine-tuned, which is true for all models leveraging pre-trained embeddings and could additionally improve the models performance \cite{thielmann2022human, bianchi2020pre}.

For LDA, ProdLDA, NeuraLLDA, ETM and CTM, we optimize over various hyperparameters with Bayesian optimization as provided by the OCTIS package \cite{terragni2021octis}. We use model perplexity, measured based on the evidence lower bound of a validation sample of documents, as the objective function  in order to not rely on metrics, such as NPMI coherence or WESS, that measure either cohesion or separation of topics. LDA is optimized over the parameters of the two symmetric Dirichlet priors on the topic-specific word distribution and the document-specific topic distribution. For ProdLDA, NeuralLDA and CTM, the learning rate parameter, as well as the number of layers and the number of neurons per layer in the inference network are considered. Finally, for ETM, we tune the learning rate, the number of hidden units in the encoder and the embedding size. 

Since BERTopic and Top2Vec are highly insensitive to different hyperparameter settings of the underlying HDBSCAN algorithm and also do not provide a way to measure the (marginal) likelihood of data, we choose the default hyperparameters for those models. While finding the optimal hyperparameters for these models might improve their performances compared to the models where we implemented hyperparameter tuning, the same is true for CBTM.

\begin{table*}[htb]
  \centering
    \begin{tabularx}{\textwidth}{l|X}
    \toprule
    Topic    & Words \\
    \hline
    \midrule
    1        & game, league, player, play, baseball, sport, pitch, hockey, team, batf \\
    2        & application, program, software, workstation, code, window, file, programming, print, tool \\
    3        & bullet, firearm, weapon, attack, shoot, kill, action, armed, protect, protection \\
    4        & homosexual, homosexuality, sexual, insist, reject, accept, morality, contrary, disagree, oppose \\
    5        & machine, chip, circuit, electronic, hardware, equipment, device, computer, workstation, processor \\
    6        & vehicle, auto, engine, rear, tire, driver, truck, motor, wheel, bike \\
    7        & israeli, conflict, oppose, attack, peace, struggle, arab, turkish, armenian, kill \\
    8        & action, consideration, complain, oppose, bother, rule, issue, policy, insist, accept \\
    9        & complain, respond, response, consideration, suggestion, idea, bother, challenge, influence, accept \\
    10       & orbit, satellite, solar, planet, shuttle, mission, earth, rocket, moon, plane \\
    11       & mailing, mail, send, email, contact, message, telephone, address, customer, request \\
    12       & printer, print, font, format, digital, make, manufacture, manufacturer, machine, workstation \\
    13       & sell, sale, purchase, offer, brand, customer, supply, vendor, deal, price \\
    14       & send, inform, publish, message, newsgroup, reader, mailing, post, topic, mail \\
    15       & lose, result, score, loss, beat, challenge, division, note, gain, fall \\
    16       & belief, faith, doctrine, accept, truth, religion, notion, religious, trust, interpretation \\
    17       & hardware, computer, device, drive, machine, monitor, electronic, chip, shareware, modem \\
    18       & patient, complain, care, affect, effect, issue, treat, suffer, response, treatment \\
    19       & interpretation, truth, assert, argue, claim, consideration, logic, insist, complain, belief \\
    20       & secure, encryption, security, encrypt, privacy, protect, protection, scheme, enforcement, access \\
    \end{tabularx}
  \caption{The CBTM model fit on the 20 Newsgroups dataset. The reference corpus is expanded with the brown corpus taken from the nltk package \cite{bird2009natural}.}
\end{table*}%

\end{document}